\documentclass[10pt,twocolumn,letterpaper]{article}

\usepackage[pagenumbers]{cvpr} 

\usepackage[dvipsnames]{xcolor}

\definecolor{cvprblue}{rgb}{0.21,0.49,0.74}
\usepackage[pagebackref,breaklinks,colorlinks,citecolor=cvprblue]{hyperref}
\usepackage{dsfont}
\usepackage{float}
\def\paperID{8} 
\def\confName{CVPR}
\def\confYear{2024}

\title{Self-Supervised Backbone Framework for Diverse Agricultural Vision Tasks}

\author{Sudhir Sornapudi, Rajhans Singh \\
Corteva Agriscience, Indianapolis, USA\\
{\tt\small sudhir.sornapudi@corteva.com, rajhans.singh@corteva.com}
}

\begin{document}
\maketitle

\begin{abstract}
   Computer vision in agriculture is game-changing with its ability to transform farming into a data-driven, precise, and sustainable industry. Deep learning has empowered agriculture vision to analyze vast, complex visual data, but heavily rely on the availability of large annotated datasets. This remains a bottleneck as manual labeling is error-prone, time-consuming, and expensive. The lack of efficient labeling approaches inspired us to consider self-supervised learning as a paradigm shift, learning meaningful feature representations from raw agricultural image data. In this work, we explore how self-supervised representation learning unlocks the potential applicability to diverse agriculture vision tasks by eliminating the need for large-scale annotated datasets. We propose a lightweight framework utilizing SimCLR, a contrastive learning approach, to pre-train a ResNet-50 backbone on a large, unannotated dataset of real-world agriculture field images.  Our experimental analysis and results indicate that the model learns robust features applicable to a broad range of downstream agriculture tasks discussed in the paper. Additionally, the reduced reliance on annotated data makes our approach more cost-effective and accessible, paving the way for broader adoption of computer vision in agriculture.
\end{abstract}
\section{Introduction}
\label{sec:intro}


Computer vision, coupled with deep learning models \cite{he2016deep, dosovitskiy2020image,he2017mask, chen2017deeplab}, has ushered in a new era of transformation for the agricultural sector. These vision models have become integral to numerous aspects of agricultural practices, including crop phenotyping \cite{shakoor2017high, aich2018deepwheat, yang2020near}, monitoring \cite{nguyen2020monitoring, kussul2017deep}, precision agriculture \cite{zheng2019cropdeep, kashyap2021towards}, agricultural robotics \cite{ciarfuglia2022pseudo, saleem2021automation}, and weed control \cite{yu2019weed}. These applications encompass everything from biological cell images to vast satellite imagery, showcasing the extensive and diverse data generated within the agricultural domain. However, the performance of deep learning models depends on the availability of large high-quality labeled datasets for training. Given the expensive and time-consuming nature of manual labeling, true potential of these datasets remains largely untapped. As agricultural datasets grow in size and complexity, the challenge of manual annotation presents a significant obstacle to unlocking their value.

\begin{figure}
  \vspace{-0.2in}
    \centering
    \includegraphics[width=0.90\linewidth]{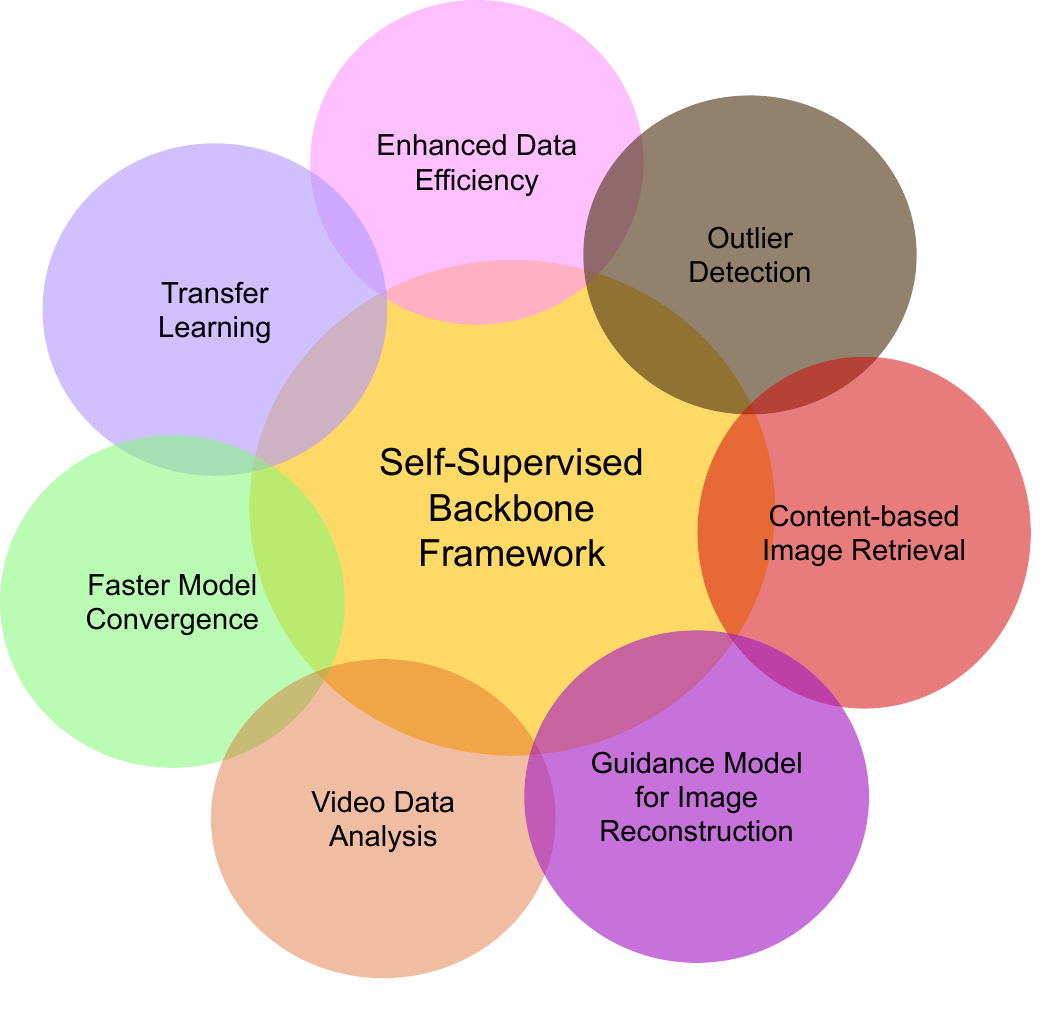}
      \vspace{-0.15in}
    \caption{Wide range of applications based on representations learned from self-supervised modeling.}
    \label{fig:benefits}
      \vspace{-0.15in}
\end{figure}

The advent of self-supervised learning (SSL) approaches \cite{jing2020self, wang2021understanding}, particularly pretraining steps, has transformed the AI landscape by enabling learning from voluminous unlabeled datasets, thereby minimizing dependency on manual labeling. This pretraining step has significantly improved performance across diverse tasks such as classification \cite{he2022masked}, segmentation \cite{araslanov2021self}, and detection \cite{wang2021dense}. By employing self-supervised techniques, models can learn generic visual representations encapsulating inherent data structures and patterns. Once pretrained, these models serve as a backbone for various tasks. Moreover, leveraging self-supervised learning with datasets from similar domains has proven to improve transfer learning capabilities as compared to data from other domains \cite{ogidi2023benchmarking}. However, publicly available large datasets from the internet often contain redundant data unrelated to agricultural tasks. 
Hence, in this paper, we investigate contrastive learning \cite{NEURIPS2020_fcbc95cc}, a self-supervised approach, utilizing large agricultural datasets for various tasks as shown in Figure \ref{fig:benefits}.

While our approach isn't the first application of self-supervised learning in the agricultural domain, notable efforts like Güldenring et al.'s \cite{GULDENRING2021106510} self-supervised contrastive learning on agriculture images and Ogidi et al.'s \cite{ogidi2023benchmarking} benchmarking of self-supervised techniques in agricultural phenotyping have set important precedents. However, these works focus on a self-supervised approach tailored for transfer learning. Despite previous endeavors, these works overlook certain advantageous aspects of the derived feature representations from self-supervision. Our paper aims to emphasize these potential benefits and applications through various empirical analyses.

Ogidi et al. \cite{ogidi2023benchmarking} demonstrated empirically that, in many instances, leveraging a domain-specific yet diverse pretraining dataset yields superior performance in transfer learning tasks. Thus, we curated a large dataset comprising agricultural field imagery captured via mobile phones and satellites for this paper. Encompassing diverse scenarios such as different crop stages, harvested and unharvested fields, and ranging from close-up leaf images to expansive field views, our dataset aims to provide a rich and varied resource for advancing agricultural computer vision research.

This work investigates how self-supervised representation learning can empower the agricultural sector by unlocking the value of unlabeled data. We explore the transformative potential of the powerful feature representations from agriculture images, leading to improved efficiency and performance across various downstream tasks. In particular, we explore the following applications and benefits:

\begin{itemize}
\item \textbf{Enhanced data efficiency:} SSL pretrained model features facilitate insightful interpretation of raw data and help in validating and balancing training dataset diversity. Moreover, fine-tuning with just $1\%$ of labeled in-domain data achieves an impressive $80.2\%$ accuracy, significantly reducing labeling requirements.
\item \textbf{Transfer learning:} Through experiments covering classification, detection, and segmentation tasks, we demonstrate the benefit of using self-supervised pretrained models over fully supervised models trained from scratch, particularly in scenarios with limited labeled training data. For these experiments, we use publicly available datasets including rice leaf disease identification \cite{Wang_2023_CVPR}, PlantVillage \cite{mohanty2016using}, and MinneApple \cite{Hani_2020}.
\item \textbf{Faster model convergence:} Experimental results from finetuning for downstream tasks demonstrate that utilizing SSL pretrained models leads to a faster training convergence rate.
\item  \textbf{Outlier detection:} Using representations from the SSL pretrained model with clustering analysis enables effective identification of anomalies in agricultural data.
\item \textbf{Content-based image retrieval at scale:} We created PixelAffinity, a web-based tool for content-based image retrieval, harnessing SSL pretrained model features to efficiently search for similar images within vast agroecology databases.
\item \textbf{Video data prepossessing:} Utilizing features extracted from SSL pretrained models on video data enables the identification of redundant frames and the selection of relevant frames tailored to specific applications.
\item \textbf{Guidance model for Image reconstruction/editing :} By employing SSL pretrained models as guidance networks, we can effectively reconstruct and synthesize new images or perform image editing tasks based on given input images, using diffusion or GAN models.

\end{itemize}

\section{Related Work}
\label{sec:relatedwork}
\begin{figure*}[ht]
  \vspace{-0.2in}
    \centering
    \includegraphics[width=0.85\linewidth]{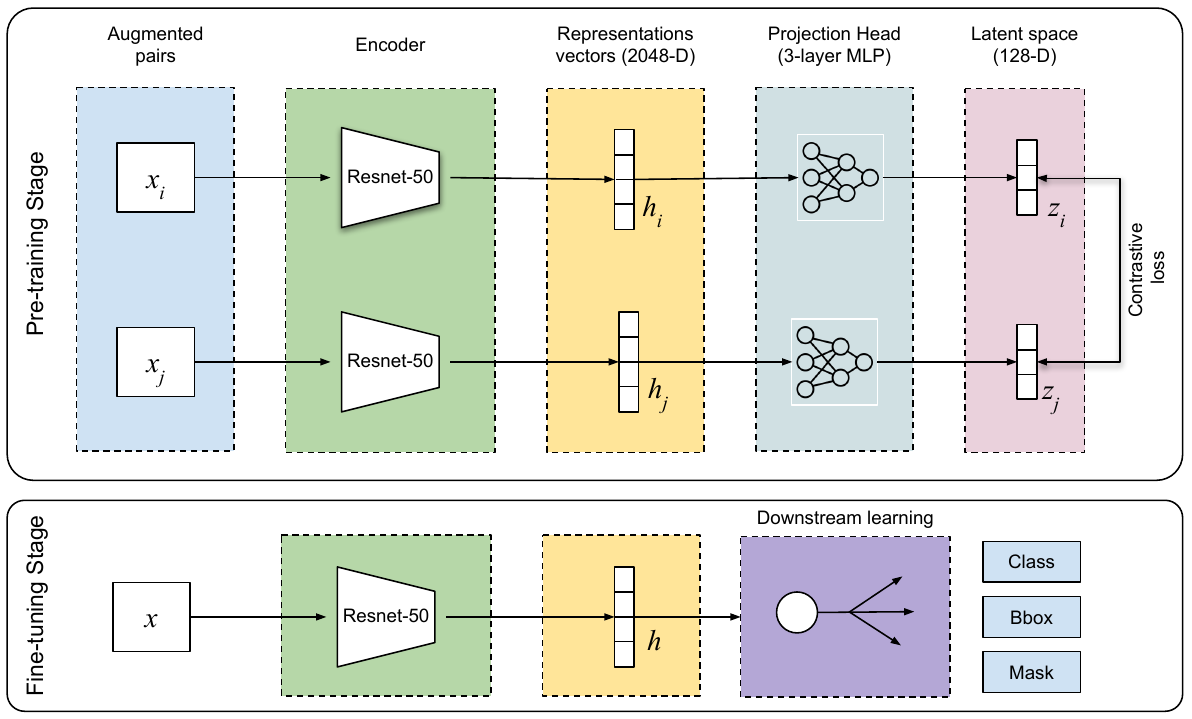}
      \vspace{-0.1in}
    \caption{Illustration of the two-stage process: self-supervised pre-training by contrasting different views of the same image and label-light supervised fine-tuning for downstream tasks.}
    \label{fig:setup}
      \vspace{-0.1in}
\end{figure*}
\noindent\textbf{Deep learning in agriculture:}
Deep learning based approach has proven highly successful in various vision-based tasks such as classification, object detection, and semantic segmentation. Its application in agriculture has yielded significant advancements across diverse classification tasks such as rice leaf disease classification \cite{Wang_2023_CVPR}, plant disease classification \cite{mohanty2016using}, sorghum classification \cite{ren2021multi}, and many more. Segmentation tasks tackle a vast range of challenges such as identifying Aphid Clusters \cite{rahman2023real}, Mushroom Segmentation \cite{retsinas2023mushroom}, and instance segmentation of lettuce \cite{wang20223d}. Additionally, detection tasks have been pursued with initiatives like Sorghum Panicle Detection \cite{cai2022high}, pollination monitoring \cite{ratnayake2021towards}, and Oil Palm Tree Detection \cite{wu2020cross}, among others. However, the performance of deep learning models for these tasks in agriculture is highly dependent on the availability of high-quality labeled datasets. In our work, we explore a self-supervised approach to reduce the burden of labeled data for training.

\noindent\textbf{Transfer learning and self-supervision in agriculture:} Due to the limited availability of labeled training data in agriculture, transfer learning and self-supervision can play pivotal roles.
For example \cite{Wang_2023_CVPR} for rice leaf disease and \cite{olsen2019deepweeds} for weed species classification use ImageNet \cite{deng2009imagenet} pretrained models, weed detection \cite{olsen2019deepweeds} and apple detection \cite{Hani_2020} use Coco \cite{lin2014microsoft} pretrained model. These publicly available large datasets for pretraining often differ significantly from the agricultural domain, underscoring the importance of leveraging pretrained models from similar domains for improved performance.
Self-supervised approaches are popular in general vision tasks; however, their application to agriculture tasks has been relatively less explored. For instance, \cite{ogidi2023benchmarking} utilizes a self-supervised approach for plant phenotyping, while \cite{yang2020self} applies it to fine-grained tomato disease classification. Despite these initial efforts, we believe that the self-supervised approach holds far greater potential than previously explored. Our paper aims to highlight these potential benefits backed by concrete evidence.

\section{Methodology}
\label{sec:formatting}

Representation learning is concerned with training machine learning algorithms to learn useful features. Deep neural networks can be considered representation learning models that typically encode information (viz. images) that is projected into a different subspace. Our approach leverages self-supervised learning in a two-stage process, as shown in Figure \ref{fig:setup}. In the pretraining stage, we employ a large, unlabeled dataset and a contrastive pretext task to train our model. This builds a foundation of generalizable feature representations. We then freeze specific layers to preserve this learned knowledge. Subsequently, in the fine-tuning stage, we adapt the pre-trained model to the specific downstream task using a smaller, labeled dataset. By strategically unfreezing relevant layers, we leverage the pre-trained weights as a starting point while simultaneously refining the model for our task objective. This two-stage approach aims to harness the power of large-scale unlabeled data for improved performance on our target task.

\subsection{Pre-training stage}

For simplicity, we adopted the SimCLR framework \cite{NEURIPS2020_fcbc95cc} that utilizes a self-supervised contrastive learning approach to learn high-quality feature representations from completely unlabeled data. Figure \ref{fig:setup} illustrates the SimCLR framework. This is a special case of unsupervised learning in which we create a pseudo-task to generate pseudo-labels, enabling pre-training in a supervised manner. This is accomplished by contrastive comparison of image pairs. A large minibatch of $N$ images is randomly augmented twice to get two different views. This results in \(2N\) data points. Given a positive pair forming from differently augmented views of the same image, the rest \(2(N-1)\) augmented views are treated as negative pairs. This way the negative pairs were implicitly sampled.
The resulting augmented pairs are passed through the same convolutional neural network (CNN). For our study, we chose ResNet-50 ($\times 1$) as the backbone CNN architecture to extract the hidden representation and eventually maximize the agreement between the two embedding vectors. This indirectly means minimizing the contrastive loss. In other words, we want to maximize the similarity between the positive pairs and minimize the similarity between the negative pairs.
The representations generated from the backbone network empirically do not generate the best results, so Chen et. al. \cite{NEURIPS2020_fcbc95cc} introduced a second projection network that generates the optimal latent space embeddings (\(z\)) to apply the loss function. We used a 3-layer multi-layer perceptron projection head to project the representation to 128-dimensional embedding. These embeddings from the projection network are only used for optimizing the training while minimizing the contrastive loss function (\(L\)). The high-quality 2048-dimensional feature representations (\(h\)) from the backbone encoder network are retained for the downstream tasks. 

For a given $x_i$, the pretext task aims to identify its positive counter part $x_j$ in $\{x_k\}_{k\neq i}$, where $\{x_k\}$ is $k$-th mini-batch set including positive pairs $x_i$ and $x_j$. The contrastive cost function \cite{NIPS2016_6b180037} for a positive pair of examples in a minibatch is defined as 
\begin{equation}
\label{eq:e}
\begin{aligned}
l_{ij} = -log\frac{exp(\mathbf{z_{i}}^T\mathbf{z_{j}}/\tau)}{\sum_{k\neq i}exp(\mathbf{z_{i}}^T\mathbf{z_{k}}/\tau)},
\end{aligned}
\end{equation}
where $\tau$ denotes a temperature parameter, and the final loss (\(L\)) is computed across all positive pairs.

We choose random crops with resize, random flip, and color jitter as data augmentations. The loss ($\tau$ = 0.1) is optimized using LARS \cite{you2017large} with a learning rate of $0.075$ and weight decay of $1e^{-4}$. The pre-training was conducted at a batch size of $2048$ for $100$ epochs with $128$ GB Cloud TPU using $8$ cores.

\subsection{Fine-tuning stage} 
After the pretraining, the model serves as a backbone for various downstream tasks such as classification, segmentation, feature visualization, and analysis. For instance, in smaller classification datasets, we fine-tune the pretrained model by appending classification heads and fine-tuning the appended MLP layers, while freezing the rest of the model. Alternatively, we can fine-tune the entire network parameters with a very small learning rate. For classification fine-tuning, we employ the Adam optimizer with a learning rate of $1e^{-5}$, trained for $100$ epochs (or $200$ for smaller datasets) with a batch size of $64$ on a single NVIDIA T4 GPU with multiclass cross-entropy loss.
In detection and segmentation tasks, we utilize the SSL pretrained ResNet-50 model as a backbone for the Mask R-CNN model \cite{he2017mask}, implemented in TensorFlow's official library. This model is trained with classification, bounding box regression, and instance segmentation loss functions. We fine-tuned this Mask R-CNN model with an image size of $640\times640$ and a batch size of $8$ on two A100 GPUs for $100k$ iterations.


\section{Experimental Analysis}
\label{sec:experiment}
This section introduces the datasets, benefits and applications of self-supervised representations, and discussion. The performance of the model is evaluated through a series of applications, and the experimental results are quantitatively and qualitatively analyzed. Finally, the effectiveness of self-supervised representation learning in digital agriculture is verified.

\subsection{Dataset}

In our work, we utilized unlabeled Corteva's real-world agriculture field data for self-supervised pretraining and publicly available datasets like PlantVillage, Rice Leaf Disease, and Minneapple for supervised fine-tuning to demonstrate robust downstream learning.

\begin{figure}
  \vspace{-0.15in}
    \centering
      \vspace{-0.15in}
    \includegraphics[width=0.94\linewidth]{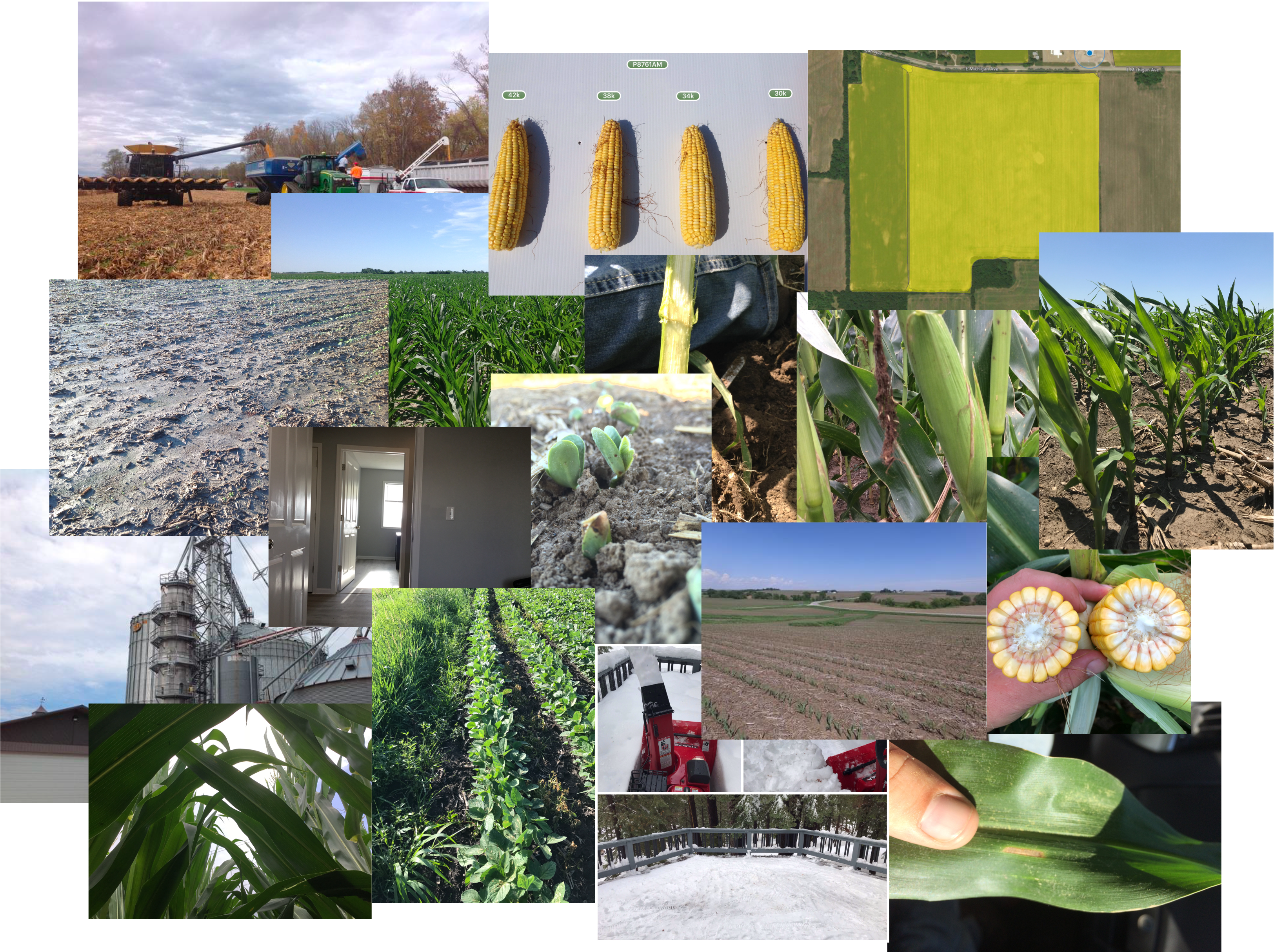}
    \caption{Corteva real-world unlabeled agriculture image data.}
    \label{fig:ag_data}
      \vspace{-0.15in}
\end{figure}

\noindent\textbf{Corteva Dataset.} 
This is an unstructured, unlabeled raw agriculture on-field and off-field image dataset captured with digital cameras, smartphones, drones, satellites, and screenshots at varying resolutions (Figure \ref{fig:ag_data}). We collected a total of $776377$ images and used the entire data for the pretraining stage to learn high-quality visual representations. We held $1\%$ of the data (i.e., $7763$ images) for finetuning and an additional $1864$ images as hold-out-test. These were labeled for downstream classification tasks. We categorized this subset of data into 8 different classes: Corn field, Corn cob, Corn leaf, Soy field, Satellite view, harvested field, Unknown field, and Other (screenshots, buildings, vehicles, manufacturing plants, etc.). This is a proprietary dataset, and will not be released to the public.

\noindent\textbf{PlantVillage Dataset.}
The PlantVillage dataset \cite{mohanty2016using} is a publicly available leaf disease dataset. The images were captured at experimental research stations in association with the Land Grant Universities. The leaves were removed from the plant and imaged using Sony DSC 20.2 megapixel digital camera. 
We used a subset of this dataset available on Kaggle\footnote{\url{https://www.kaggle.com/datasets/emmarex/plantdisease/data}}. It consists of $20.6$K healthy and unhealthy leaf images. These images were labeled by expert plant pathologists and were divided into 15 categories based on 3 crop species (bell pepper, potato, and tomato) and various diseases (bacterial, early and late blight, leaf mold, spots, 2 viral and 1 caused by mite).
We partitioned this dataset into training and test sets using an $80-20\%$ split, ensuring a uniform selection from each class. We used this dataset to demonstrate the transfer learning capability of our pre-trained and fine-tuned models.

\noindent\textbf{Rice Leaf Disease Identification Dataset.}
This dataset \cite{Wang_2023_CVPR} was aggregated from publicly available rice leaf datasets and the internet, and annotated by researchers under the guidance of plant protection experts. The dataset consists of $4523$ rice leaf disease images and is sorted into seven classes (healthy, hispa, brown spot, leaf blast, bacterial blight, bacterial leaf streak, and sheath blight). The authors released a subset of $3353$ annotated images on Kaggle, a platform for data science competitions. We split the dataset $80-20\%$ into training and test sets, uniformly sampling from all classes, and we used this subset of images for our downstream classification task.

\noindent\textbf{MinneApple Dataset.}
The dataset \cite{Hani_2020} was created by the University of Minnesota researchers to create a unified dataset to advance the state-of-the-art in fruit counting, detection, and segmentation in orchard environments. These are a variety of high-resolution red and green apple fruit on tree images. The dataset contains $1000$ images with over $41000$ annotated object instances. The annotations span over patch-based counting of clustered fruits, marked bounding boxes, and masks drawn over each apple instance. We use this dataset to study the transferability of the base ResNet-50 model for detection and segmentation tasks.

\noindent\textbf{Sentinel-2 satellite data.}
Sentinel-2 data is part of the Copernicus Sentinel-2 mission \cite{sentinel2}, which aims to provide free and open access to Earth observation data captured by two polar-orbiting satellites. Although Sentinel-2 is a multi-spectral instrument, we focused on retrieving data that contain RGB bands at 10 meters spatial resolution from 10 different field locations around the globe for our experimental purposes, specifically from Level-2A which offers atmospherically corrected surface reflectance images. We prepared a $338$ agriculture images dataset with $256\times256$ resolution, containing regions with green vegetation, no vegetation, and water, with and without cloud shadow, and a few medium and high cloud cover images. We use this dataset to observe the outlier detection capability of the proposed model, where images with cloud cover are outliers.

\subsection{Benefits and Applications}

\begin{figure}
  \vspace{-0.25in}
    \centering
    \includegraphics[width=0.90\linewidth]{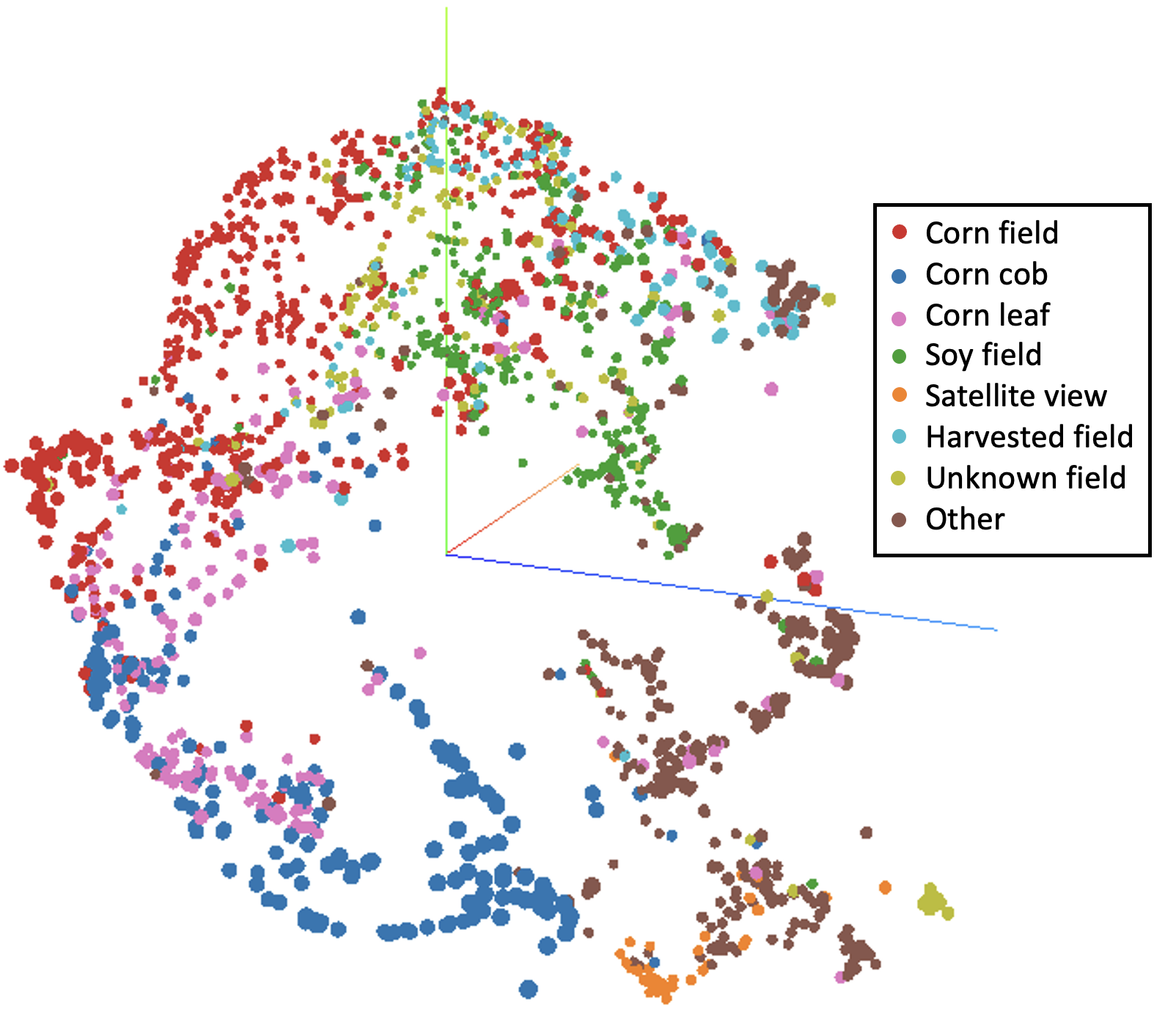}
      \vspace{-0.10in}
    \caption{3D point cloud visualization of hold-out Corteva test data feature representations with color-coded ground-truth labels.}
    \label{fig:tb-pointcloud}
      \vspace{-0.15in}
\end{figure}

\noindent\textbf{Enhanced Data Efficiency}\newline
We begin by examining feature visualization of the SSL pretrained model as a means to enhance data efficiency. In Figure \ref{fig:tb-pointcloud}, we present a 3D point cloud comprising $1864$ labeled test set, where each point corresponds to a 2048-dimensional feature embedding vector representing its respective input image (plotted in a 3D space using UMAP dimension reduction algorithm \cite{mcinnes2020umap}). The figure also shows the ground-truth labels of each point as color code. The cluster formations, indicated by ground truth labels, emphasize the model's ability to learn robust feature representation, thereby serving as an effective automated labeling tool. This can be extended to observe the training data points of downstream tasks and understand the cluster formations to evaluate the diversity and imbalance in the training data. Further, we leverage the pretrained model for fine-tuning on a $1\%$ of labeled data from our Corteva dataset across eight classes. Subsequently, we evaluated the model on $1864$ held-out test images, achieving an impressive accuracy of $80.2\%$ accuracy despite utilizing only small subset of labeled information during training.
\newline

\noindent\textbf{Transfer Learning}\newline
For transfer learning, we used two datasets for classification and one dataset for detection and instance segmentation.  For the classification tasks, we adopted a fine-tuning approach with pre-trained models and contrasted the results against models trained from scratch and fine-tuned from ImageNet weights on both datasets. Additionally, we conducted experiments using only $25\%$ and $10\%$ of the training datasets and evaluated their performance on the entire training set. Tables \ref{table:rice_disease_accuracy} and \ref{table:plantvillage_accuracy} show the accuracy of various models on the rice leaf disease and plant village datasets respectively. In rice disease classification, fine-tuning the entire parameter set yielded a $1.2\%$ improvement when using $100\%$ of the training data, and a notable $17\%$ and $4\%$ improvement with only $10\%$ of the data, compared to models trained from scratch and fine-tuning ImageNet, respectively. Similarly, on the plant village dataset, while performance remained almost the same with $100\%$ of the training data, a substantial $7\%$ and $3\%$ respective improvement was observed when utilizing only $10\%$ of the data. Interestingly, fine-tuning only the last layer resulted in inferior performance compared to models trained from scratch. These tables indicate that in scenarios with limited training data, the fine-tuning approach significantly outperforms models trained from scratch. 

\begin{figure}[]
     \centering
     \vspace{-0.2in}
         \includegraphics[width=0.85\columnwidth]{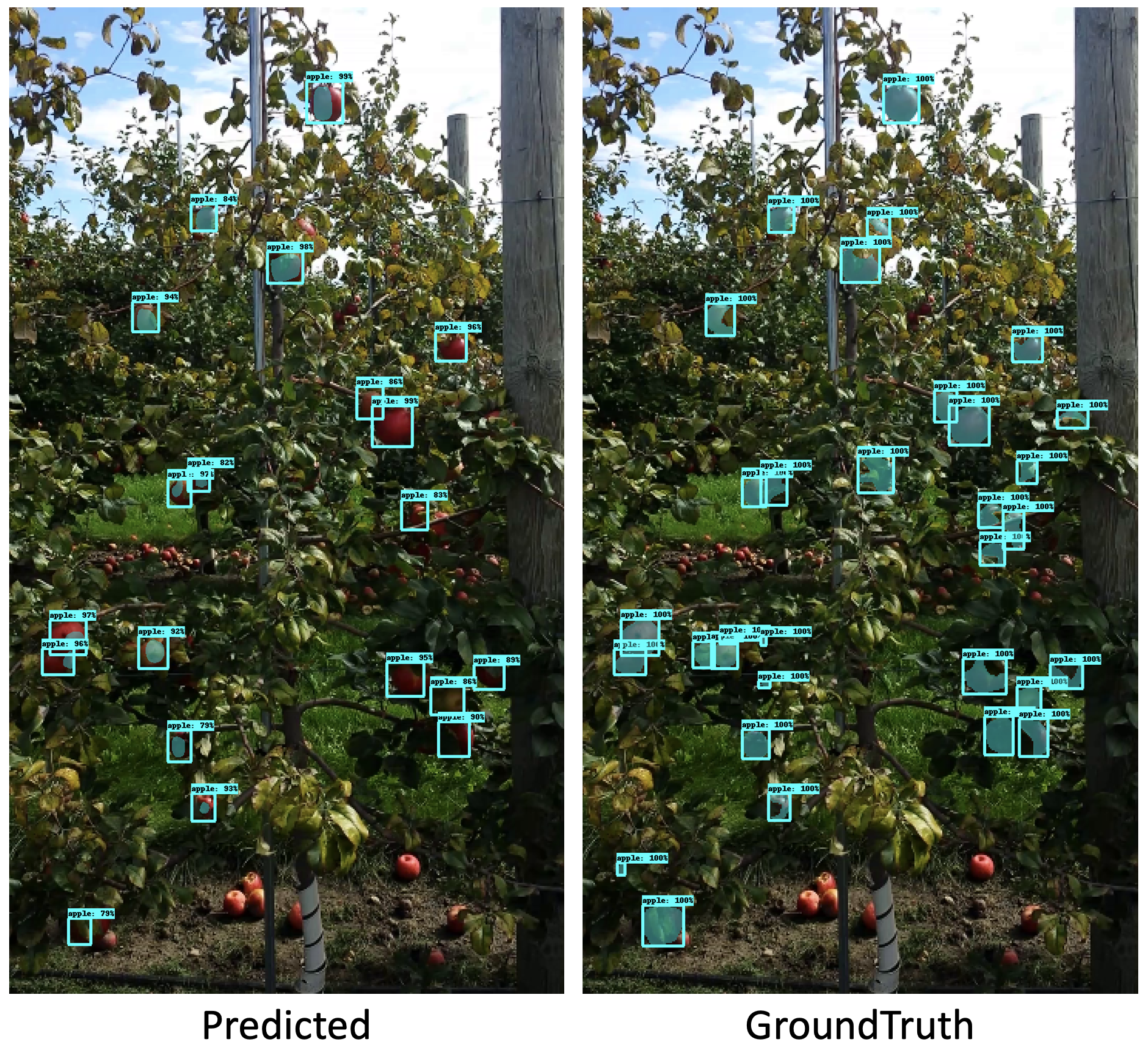}
         \vspace{-0.10in}
        \caption{Instance segmentation and detection result from Mask R-CNN model. Finetuned model from Self-supervised pretrained ResNet-50 weights (left) and ground-truth annotation (right).}
        \label{fig:rcnn_detection}
        \vspace{-0.10in}
\end{figure}

Table \ref{table:minneapple_detection} presents the mean Average Precision (mAP) results for object detection on the minneapple dataset using the Mask R-CNN architecture. The experiments involve training with three different setups: one starting from scratch, and the other two utilizing pretrained weights from ImageNet and SSL pretrained weights from the Corteva dataset. The table shows nearly equal performance using pretrained backbone weights compared to training from scratch. However, the backbone model pretrained on the Corteva dataset with self-supervision outperforms the ImageNet pretrained model, underscoring the significance of using domain-specific data for pretraining backbone models. Figure \ref{fig:rcnn_detection} shows the visualization of detection results alongside ground truth, obtained from the Mask R-CNN utilizing our SSL pretrained Corteva backbone model.
\newline
\begin{table}[ht!]
\vspace{-0.1in}
\caption{Performance comparison of finetuning and training from the scratch approach on rice leaf disease classification using ResNet-50 model.}
\label{table:rice_disease_accuracy}
  \vspace{-0.1in}
\centering
\begin{small}
\begin{tabular}{cccc}
\toprule
\multicolumn{1}{c}{\textbf{Training}} & \multicolumn{1}{c}{\textbf{Training From}}&  \multicolumn{1}{c}{\textbf{Finetuning}} &  \multicolumn{1}{c}{\textbf{Finetuning}}\\
\textbf{Data}& \textbf{Scratch}& \textbf{ImageNet}& \textbf{\textbf{SSL Pretrained}}\\
\midrule
100\%& 95.99& 96.12 &  \textbf{97.18}\\
25\% & 79.49& 85.88 & \textbf{90.49}\\
10\% &63.00& 76.37 &  \textbf{80.83}\\
 \bottomrule
\end{tabular}
\end{small}
\end{table}

\begin{table}[ht!]
\vspace{-0.1in}
\caption{Performance comparison of finetuning and training from scratch approach on plant village classification using ResNet-50 model.}
\label{table:plantvillage_accuracy}
  \vspace{-0.1in}
\centering
\begin{small}
\begin{tabular}{cccc}
\toprule
\multicolumn{1}{c}{\textbf{Training}} & \multicolumn{1}{c}{\textbf{Training From}}&  \multicolumn{1}{c}{\textbf{Finetuning}} &  \multicolumn{1}{c}{\textbf{Finetuning}}\\
\textbf{Data}& \textbf{Scratch} & \textbf{ImageNet}&\textbf{SSL Pretrained}\\
\midrule
100\%& 99.56 & 99.54&  \textbf{99.59}\\
25\% & 97.39 &97.77& \textbf{98.79}\\
10\% &90.30 &94.44& \textbf{97.22}\\
 \bottomrule
\end{tabular}
\end{small}
\end{table}

\begin{table}[ht!]
\vspace{-0.1in}
\caption{Performance comparison of finetuning and training from scratch approach on Minneapple object detection dataset using Mask R-CNN with ResNet-50 backbone.}
\label{table:minneapple_detection}
  \vspace{-0.1in}
\centering
\begin{small}
\begin{tabular}{cccc}
\toprule
\multicolumn{1}{c}{\textbf{Metric}} & \multicolumn{1}{c}{\textbf{Training From}}&  \multicolumn{1}{c}{\textbf{Finetuning}} &  \multicolumn{1}{c}{\textbf{Finetuning}}\\
& \textbf{Scratch} & \textbf{ImageNet}&\textbf{SSL Pretrained}\\
\midrule
\textbf{AP[.5:.95]} & 0.23 & 0.22 & \textbf{0.24}\\
\textbf{AP[.5]} & \textbf{0.60} & 0.59 & 0.59 \\
\textbf{AP[.75]} & 0.12 & 0.11 & \textbf{0.16}\\
 \bottomrule
\end{tabular}
\end{small}
\end{table}

\begin{figure}[]
\vspace{-0.2in}
  \centering
  \begin{subfigure}{0.238\textwidth}
    \includegraphics[width=\textwidth]{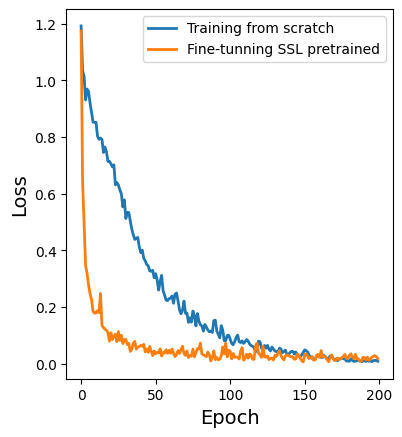}
    \caption{Rice leaf disease classification}
    \label{fig:subfig_rice_loss_converge}
  \end{subfigure}
  \begin{subfigure}{0.233\textwidth}
    \includegraphics[width=\textwidth]{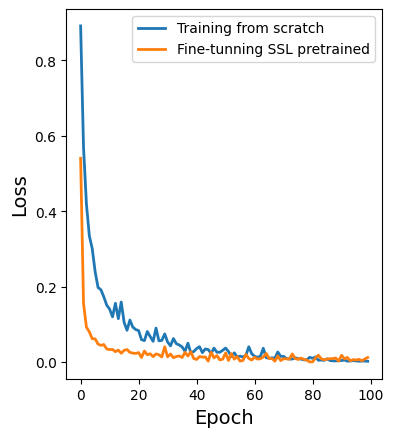}
    \caption{PlantVillage classification}
    \label{fig:subfig_plant_village_loss_converge}
  \end{subfigure}
    \vspace{-0.15in}
  \caption{Training loss convergence vs epoch for classification downstream task. The self-supervised finetuning approach leads to faster convergence compared to training from scratch.}
  \label{fig:loss_convergence}
  \vspace{-0.1in}
\end{figure}

\noindent\textbf{Faster Model Convergence}\newline
Using pretrained models for downstream tasks accelerates model convergence, helping the model learn faster and more efficiently. Figure \ref{fig:loss_convergence} shows the cross-entropy loss values plotted against training epochs for the classification tasks on the rice leaf disease and PlantVillage datasets, comparing models trained from scratch and SSL finetuned. The graphs indicate that SSL finetuned models achieve faster convergence compared to models trained from scratch. We observed comparable convergence rates and sometimes better with SSL finetuned when contrasted with ImageNet fine-tuned models. \newline

\noindent\textbf{Outlier Detection}\newline
Outlier detection, also known as anomaly detection, refers to the process of identifying patterns in data that deviate significantly from the norm or expected behavior. For instance, in agricultural contexts, anomalies could signify diseased crops, pest infestations, cloud cover, obstacles, or environmental stressors that require attention. Leveraging the feature representations learned by pretrained models can effectively differentiate between normal and anomalous patterns in agricultural data. Furthermore, this contributes to identifying data imbalance and diversifying the training data, thereby improving the model’s capacity to generalize and make precise predictions on unseen data.

\begin{figure}
    \vspace{-0.15in}
    \centering
    \includegraphics[width=0.90\linewidth]{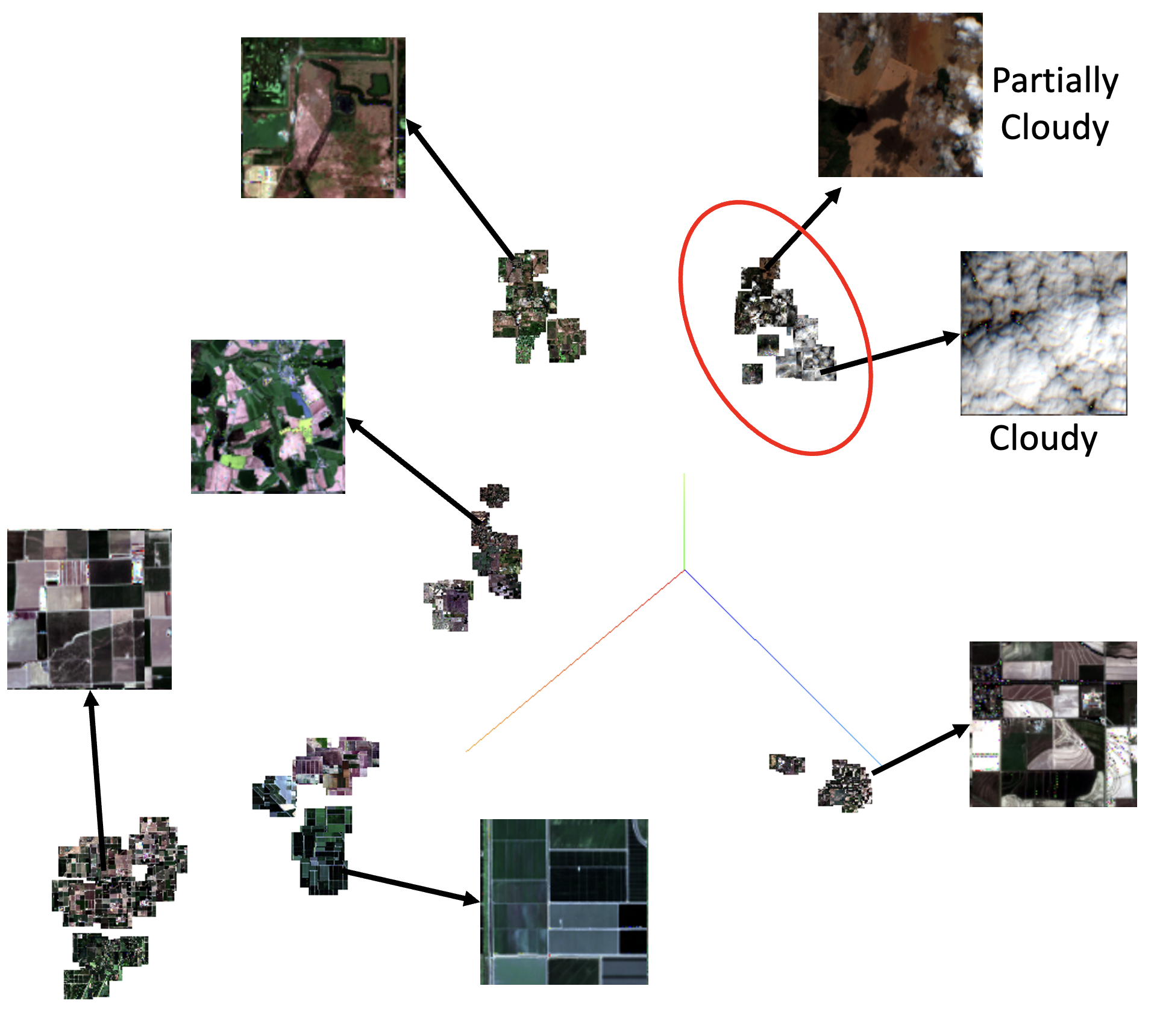}
    \vspace{-0.1in}
    \caption{Visualization of cloud cover outlier clusters in Sentinel-2 agricultural field satellite data. This visualization shows the clustering of partial or complete cloud covers (outlier marked in red) and also different landscapes, including green vegetation, no vegetation, water bodies, location, etc.}
    \label{fig:tb-cloudcover}
    \vspace{-0.05in}
\end{figure}

We illustrate this utility by employing the Sentinel-2 dataset, discerning the cloud coverage from agricultural field satellite imagery data. We utilized the SSL pretrained ResNet-50 model to derive $2048$ dimensional image representation vectors and projected them onto a 3D plane using the UMAP dimensionality reduction algorithm. The cluster formations can be observed from Figure \ref{fig:tb-cloudcover}, facilitating the identification of a minor cluster with the cloud cover outlier images from the Sentinel-2 dataset. Consequently, we can efficiently exclude undesirable cloud cover images from the high-altitude field images captured by drones, flights, and satellites.\newline

\noindent\textbf{Content-based Image Retrieval}\newline
AI applications in agriculture analysis may also encounter rare and complicated cases that are hard for normal agronomists to analyze due to their scarcity. Content-based image retrieval (CBIR) means AI’s ability to search for images similar to those retrieved from large agroecology databases. CBIR is very helpful in the agriculture analysis process because it improves the chance of making the right analysis in challenging cases by quickly finding examples of cases that are similar to available agroecology image databases that have such rare and complicated cases. In this case, similarity reflects related agroecological intrinsic feature representations rather than just visual similarities.

\begin{figure}
    \centering
    \includegraphics[width=0.9\linewidth]{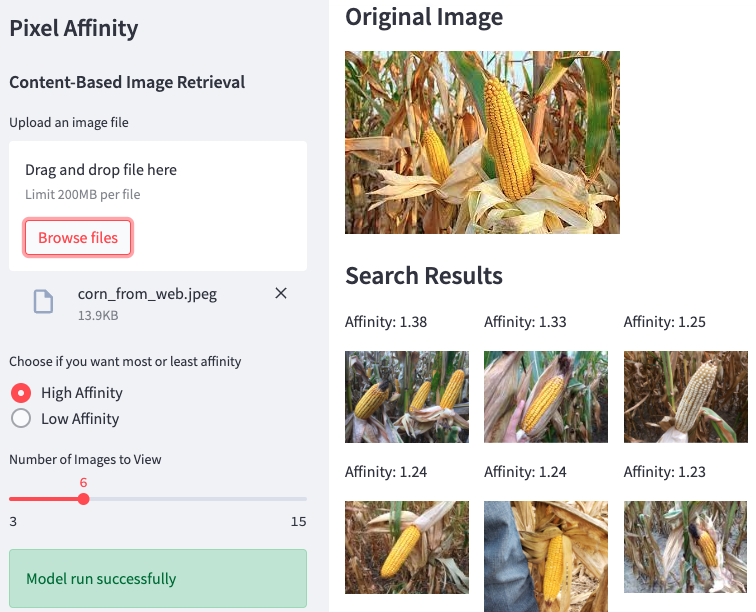}
      \vspace{-0.1in}
    \caption{Pixel Affinity web application user interface for content-based image retrieval.}
    \label{fig:pa-ui}
      \vspace{-0.15in}
\end{figure}

\begin{figure}
  \vspace{-0.1in}
    \centering
    \includegraphics[width=0.80\linewidth]{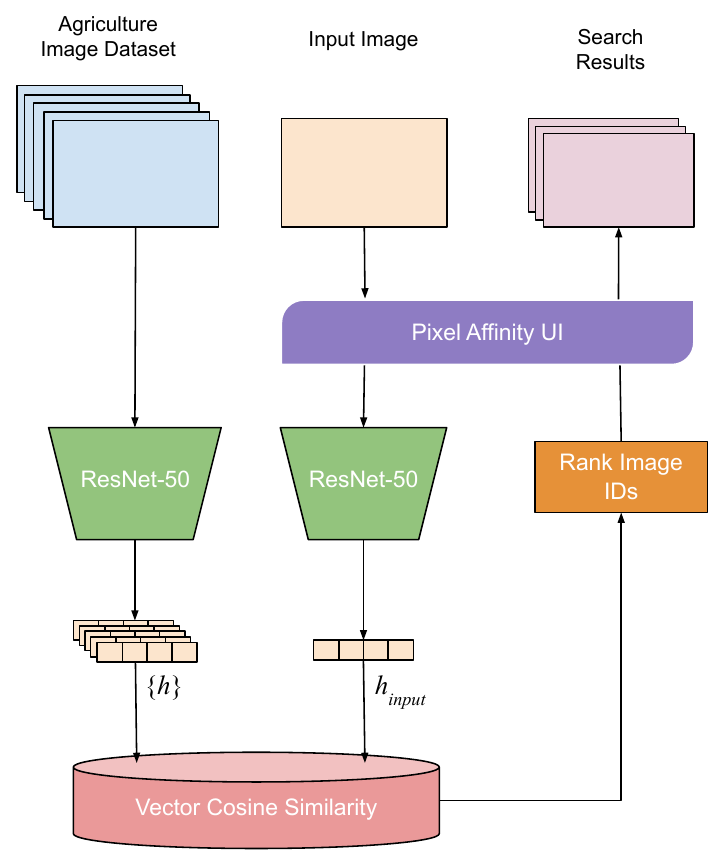}
      \vspace{-0.1in}
    \caption{Block diagram of Pixel Affinity tool for content-based image retrieval.}
    \label{fig:pa-bd}
      \vspace{-0.15in}
\end{figure}

We have designed PixelAffinity, a web-based CBIR search tool (Figure \ref{fig:pa-ui}), which leverages the proposed method to deliver accurate results. We considered the fine-tuned ResNet-50 model to create robust image representations. These representations form an offline database, allowing for efficient searches. When a user uploads an image, its representation is generated online processed by a CPU, and compared to the database using cosine similarity. This vector similarity search is performed at scale for high throughput. The closest matches are then displayed in the search results, offering users a fast and effective way to find similar images. This entire process can be visualized from Figure \ref{fig:pa-bd}.\newline

\noindent\textbf{Video Data Analysis}\newline
Video analysis can help improve productivity in agriculture from field monitoring to harvest automation. Drones equipped with cameras survey fields, providing data for precision planning like targeted fertilizer application. Robots guided by video analysis automate plant phenotyping and harvesting, minimizing waste. This translates to better decision-making, boosting productivity and efficiency across the farm.

\begin{figure}
  \vspace{-0.20in}
    \centering
    \includegraphics[width=0.90\linewidth]{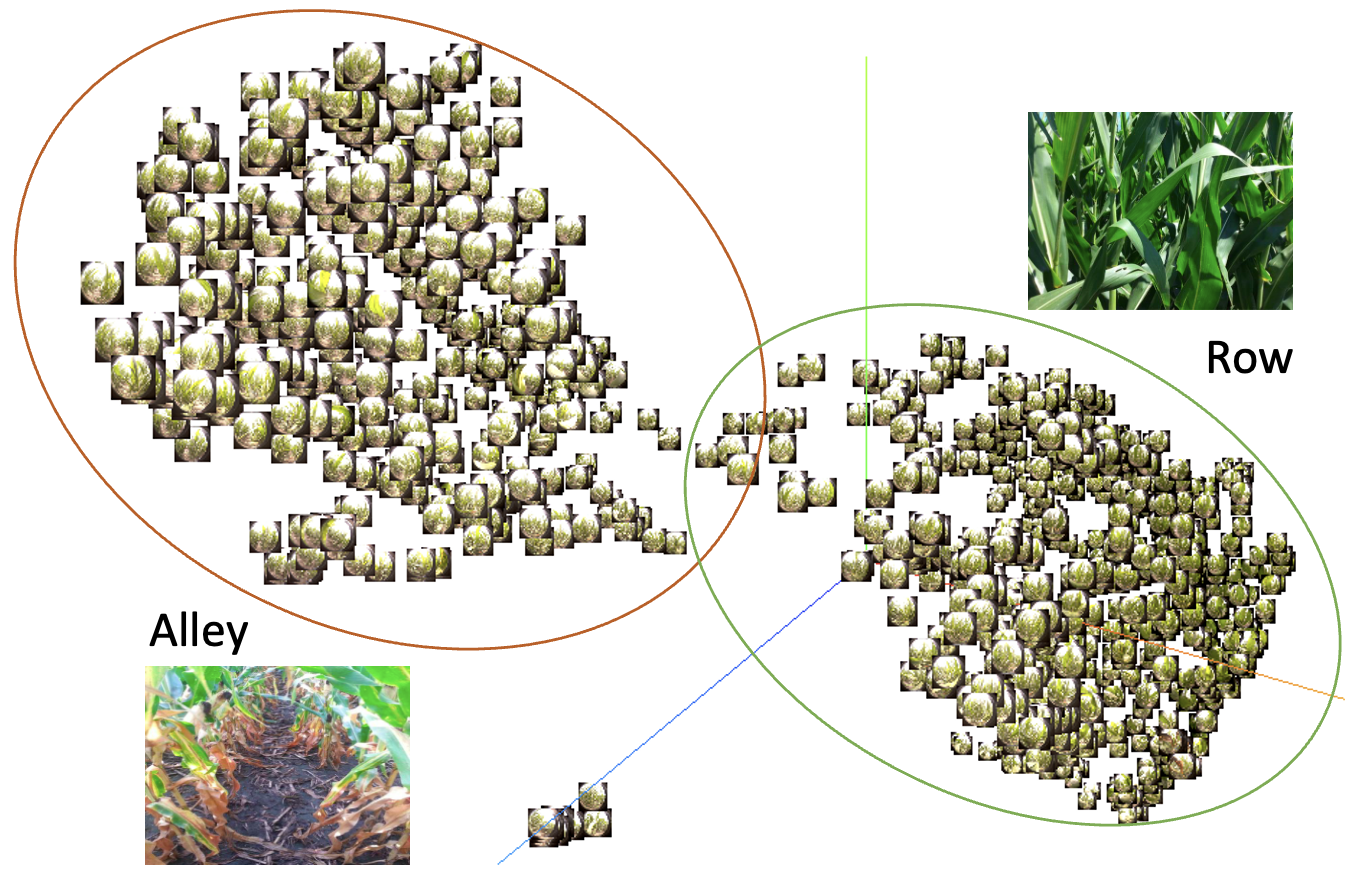}
      \vspace{-0.15in}
    \caption{Visualization of video frames as row vs alley clusters in 3D space based on its respective feature representations.}
    \label{fig:tb-frames}
      \vspace{-0.2in}
\end{figure}

Our seed research facilities conduct numerous experiments and capture on-farm video data to be processed for automated phenotyping purposes. This can be quite compute-intensive to process all the frames in a video. We studied corn field videos for corn phenotyping and found a real value by cutting video frame processing by half by separating rows and alley frames using the feature representations generated from the proposed model. Figure \ref{fig:tb-frames} shows representations plotted in a 3D space by the UMAP dimensionality reduction technique. We can observe the formation of two clusters based on rows and alleys in the frames. For phenotyping purposes, 
 breeders are interested in rows with corn plants, and alley frames were discarded cutting down the time, resources, and storage in half. \newline

\noindent\textbf{Guidance model for Image reconstruction}\newline
We observed that self-supervised training offers powerful representations from raw raw imagery. We exploit the representations and pre-trained encoder to guide the decoder for image reconstruction and image editing. Designing an effective decoder presents unique challenges, and we are investigating several promising avenues. One approach utilizes Generative Adversarial Networks (GANs) \cite{goodfellow2020generative}, where a discerning discriminator evaluates the reconstructed images. Another method employs a PixelCNN decoder \cite{van2016conditional}, iteratively predicting each pixel's value while conditioned on the representation. Additionally, we are exploring diffusion models \cite{dhariwal2021diffusion}, capable of synthesizing images guided by the extracted information. To ensure reconstructed images retain high-level visual quality, we employ perceptual loss functions like the perceptual similarity index (PSI) or style loss. This is currently a work in progress and we aim to unlock the power of self-supervised learning for accurate and visually pleasing image reconstruction.

\subsection{Discussion}
In this paper, we delve into the advantages of self-supervised learning within the agricultural domain. Our experiments demonstrate that using pretraining techniques significantly reduces the need for manual labeling while enhancing downstream task performance. This is evident from Table \ref{table:rice_disease_accuracy} and Table \ref{table:plantvillage_accuracy}, especially in the case of 10\% of training data. Notably, pretrained models using agricultural domain datasets outperform those trained on general vision datasets, highlighting the importance of domain-specific knowledge for effective representation learning. This translates to a substantial decrease in the need for manual labeling, a critical bottleneck in traditional supervised learning approaches. Moreover, leveraging feature representations from pretrained models facilitates the analysis of vast raw data, such as videos, breathing new life into previously underutilized resources, and identifying unknown outliers in the dataset. Furthermore, these representations prove effective in image retrieval tasks, streamlining the process of finding similar images across diverse databases within large agricultural organizations and fostering collaboration across teams for training data creation and knowledge exchange.

\noindent\textbf{Challenges and future work:} While pretrained models offer improved performance for downstream tasks, they still require significant computational resources for training. Additionally, the vast spectrum of agricultural tasks, ranging from microscopic images to satellite imagery, poses a significant challenge in curating large datasets for pretraining purposes. In our study, we focused solely on the ResNet-50 model; however, it is widely known that employing deeper or more advanced architectures, such as self-attention-based vision transformers \cite{dosovitskiy2020image}, could yield even better performance. Investigating these architectures and exploring recent training techniques and approaches such as DINOv2 \cite{oquab2023dinov2} and mask autoencoder \cite{he2022masked} in the agricultural domain would be of great interest. Choosing the right self-supervised pretext task (contrastive learning in our case) is crucial and challenging research problem as it affects downstream decision-making capability because pretexts rely on exploiting the structure of the data. It is a good practice to think of the properties we want in our representations to decide on a pretext task. Additionally, for specific downstream tasks like detection or segmentation, leveraging dense prediction-based self-supervised approaches \cite{wang2021dense} prove more effective than global feature-based methods. These avenues represent promising directions for future research in enhancing the efficacy of self-supervised learning within agricultural contexts.

\section{Conclusion}
\label{sec:conclusion}
This work demonstrates the significant promise of self-supervised representation learning for agricultural applications. By leveraging readily available unlabeled data, self-supervision proves to be a powerful backbone framework for significantly reducing manual labeling efforts for efficient downstream learning tasks, analyzing large-scale data through clustering, identifying anomalies, enhancing video analysis, and facilitating image search functionality within agricultural organizations. We believe that further research in this area, including exploring advanced architectures, identifying ideal self-supervised pretext tasks, tailoring approaches to specific needs, and addressing computational limitations, will unlock the full potential of self-supervised representation learning and empower agriculture to thrive in a data-driven future. Finally, our study outlines and encourages more exploration of the versatile use of self-supervised methods and their generated meaningful low-level feature representations.
{
    \small
    \bibliographystyle{ieeenat_fullname}
    \bibliography{main}
}


\end{document}